\documentclass{article}
\usepackage{ijcai21}
\usepackage{color}
\usepackage{xcolor}
\usepackage{times}
\usepackage{url}
\usepackage[hidelinks]{hyperref}
\usepackage[utf8]{inputenc}
\usepackage[small]{caption}
\usepackage{graphicx}
\usepackage{amsmath}
\usepackage{amsthm}
\usepackage{booktabs}
\usepackage{algorithm}
\usepackage{algpseudocode}
\usepackage[noend]{algcompatible}
\usepackage{amssymb}
\usepackage{multirow}
\usepackage{float}
\usepackage{booktabs}
\usepackage{adjustbox}
\usepackage{array}
\usepackage{bookmark}
\usepackage{enumitem}

\title{Dynamic Attention-based Communication-Efficient Federated Learning}

\author{
Zihan Chen$^{1,2}$\and 
Kai Fong Ernest Chong$^1$\And
Tony Q.S. Quek$^1$\\
\affiliations
$^1$Singapore University of Technology and Design\\
$^2$National University of Singapore\\
\emails
zihan\_chen@mymail.sutd.edu.sg,
\{ernest\_chong, tonyquek\}@sutd.edu.sg
}

\begin{document}

\maketitle

\begin{abstract}
Federated learning (FL) offers a solution to train a global machine learning model while still maintaining data privacy, without needing access to data stored locally at the clients. However, FL suffers performance degradation when client data distribution is non-IID, and a longer training duration to combat this degradation may not necessarily be feasible due to communication limitations. To address this challenge, we propose a new adaptive training algorithm \texttt{AdaFL}, which comprises two components: (i) an attention-based client selection mechanism for a fairer training scheme among the clients; and (ii) a dynamic fraction method to balance the trade-off between performance stability and communication efficiency. Experimental results show that our \texttt{AdaFL} algorithm outperforms the usual \texttt{FedAvg} algorithm, and can be incorporated to further improve various state-of-the-art FL algorithms, with respect to three aspects: model accuracy, performance stability, and communication efficiency. 
\end{abstract}

\section{Introduction}
\label{sec:intro}
With the proliferation of digitalized data, and with the rapid advances in deep learning, there is an ever-increasing demand to benefit from insights that can be extracted from state-of-the-art deep learning models, while still not sacrificing on data privacy \cite{geyer2017differentially,yang2019federated}. 
For effective training, these models would require a huge amount of data, which is usually distributed over a large heterogeneous network of clients who are not willing to share their private data. Federated learning (FL) promises to solve these key issues, by training a global model collaboratively over decentralized data at local clients \cite{konevcny2016federated2,mcmahan2017communication}.

The key idea of FL is that all clients keep their private data and share a global model under the coordination of a central server, which is summarized as the \texttt{FedAvg} algorithm \cite{mcmahan2017communication}. 
In a typical communication round, a fraction of the clients, selected by the server, would download the current global model and perform training on local data.  
Next, the server performs model aggregation to 
compute a new global model, based on all updated models transmitted by the selected clients. 
However, in real-world non-IID data distributions, FL in heterogeneous networks would face many statistical challenges, such as model divergence and accuracy reduction \cite{smith2017federated,li2019convergence,ghosh2019robust,karimireddy2020scaffold}.

{\bf Related work:} A theoretical analysis of the convergence of \texttt{FedAvg} in non-IID data settings is given in \cite{li2019convergence}.  
\texttt{FedProx}, a general optimization framework with robust convergence guarantees, was proposed in \cite{li2020prox} to tackle data heterogeneity by introducing a proximal term in the local optimization process. 
In \cite{karimireddy2020scaffold}, a stochastic algorithm \texttt{SCAFFOLD} was proposed to reduce the variance and gradient dissimilarity in local FL updates, which yields better communication efficiency and faster convergence. To address fairness issues in FL, $q$-Fair Federated Learning (\texttt{\emph{q}-FFL})~\cite{li2019fair} and Agnostic Federated Learning (\texttt{AFL})~\cite{mohri2019agnostic} were proposed as modified federated optimization algorithms that improve fairness in performance across clients.

In \cite{wang2019fedma}, a layer-wise Federated Matched Averaging (\texttt{FedMA}) algorithm was proposed for CNN-based and LSTM-based FL models. 
Ji \emph{et al.}~\shortcite{ji2019learning} also introduced a layer-wise attentive aggregation method for federated optimization, which has lower perplexity and communication cost for neural language models. 
Other methods have been proposed to tackle data heterogeneity and communication efficiency, from the perspective of asynchronous optimization \cite{xie2019asynchronous}, model compression \cite{konevcny2016federated2,sattler2019robust}, personalized training \cite{deng2020adaptive}, and incentive design \cite{yu2020fairness,kang2019incentive}. 

There are also several works in FL that tackle performance degradation in non-IID data settings, but for which the data privacy assumption in FL is not strictly adhered to. 
These works require a small amount of raw data to be shared either among clients \cite{zhao2018federated} or with the global server \cite{jeong2018communication}. 
\texttt{FedMix} \cite{yoon2021fedmix} avoids the direct sharing of raw data, by incorporating \textit{mixup} data augmentation \cite{zhang2018mixup} into FL, whereby clients share averaged data with each other. 
Such methods would require the exchange of raw or averaged data, which no longer guarantee strict data privacy, and which also require additional communication cost. 

In each communication round of the usual FL framework, a fixed fraction of the clients is selected, based on a fixed probability distribution. 
A good choice for this fraction is not clear.
Small fractions are widely used in existing work  \cite{mcmahan2017communication,karimireddy2020scaffold}, but when such small fractions are used, data heterogeneity inadvertently causes fluctuations in training performance and reduces the rate of convergence~\cite{li2019convergence}. 
Large fractions yield a more stable performance, as well as a slight acceleration in convergence, but at the expense of a larger communication cost~\cite{li2019convergence}. 
Hence, to obtain a stable training performance with relatively low communication cost, we shall instead consider a dynamic fraction method that captures the advantages of both small and large fractions.

The selection probability for each client is a measure of the ``importance'' of that client. Thus, in heterogeneous networks where different clients have different ``importance'', the fixed client selection probability distribution used in the usual FL is typically non-uniform. 
However, the relative contribution of each client is fluid. It depends on both the client's local model performance, and on the actual aggregated global model, so the ``importance'' of the clients may vary during training. 

With these considerations, we propose 
an attention-based adaptive federated learning algorithm,
which we shall call \texttt{AdaFL}. 
The attention mechanism in \texttt{AdaFL} serves to better capture the relative ``importance'' among the clients, by taking into account the divergence of the local updated models, relative to the global model. 
Our method gives clients who have worse models a higher chance to participate in training. 
\texttt{AdaFL} also incorporates a dynamic fraction method to balance the trade-off between small and large fractions, 
which essentially represents the trade-off between communication efficiency and performance stability.

Our contributions are summarized as follows: 
\setlist{nosep}
\begin{itemize}
    \item We introduce a method to update the client selection probability in FL, by using a distance-based attention mechanism, so as to better capture the heterogeneity of client performance. 
         
    \item To the best of our knowledge, we are the first to propose a dynamic fraction method in FL.
    We show that by increasing the fraction progressively, we can improve both the performance stability and the final model accuracy, while concurrently achieving lower communication and computation costs.
    
    \item We propose \texttt{AdaFL}, which combines both methods. We show experimentally the outperformance of \texttt{AdaFL} over \texttt{FedAvg} with respect to model accuracy, performance stability, and communication efficiency. We also show that \texttt{AdaFL} can be incorporated to enhance the performance of various state-of-the-art FL algorithms.
\end{itemize}

\section{Proposed method}
\label{sec:method}

The training of FL would typically be performed based on updates over hundreds, or even thousands of communication rounds. 
In this section, we introduce the preliminaries of FL ~\cite{mcmahan2017communication}, and show the details of our proposed algorithm within each round. 
For concreteness, we assume in this paper that FL is used to learn a global neural network model.

\begin{figure}[tb!]

  \begin{minipage}[b]{1.0\linewidth}
    \centering
    \centerline{\includegraphics[width=8.5cm]{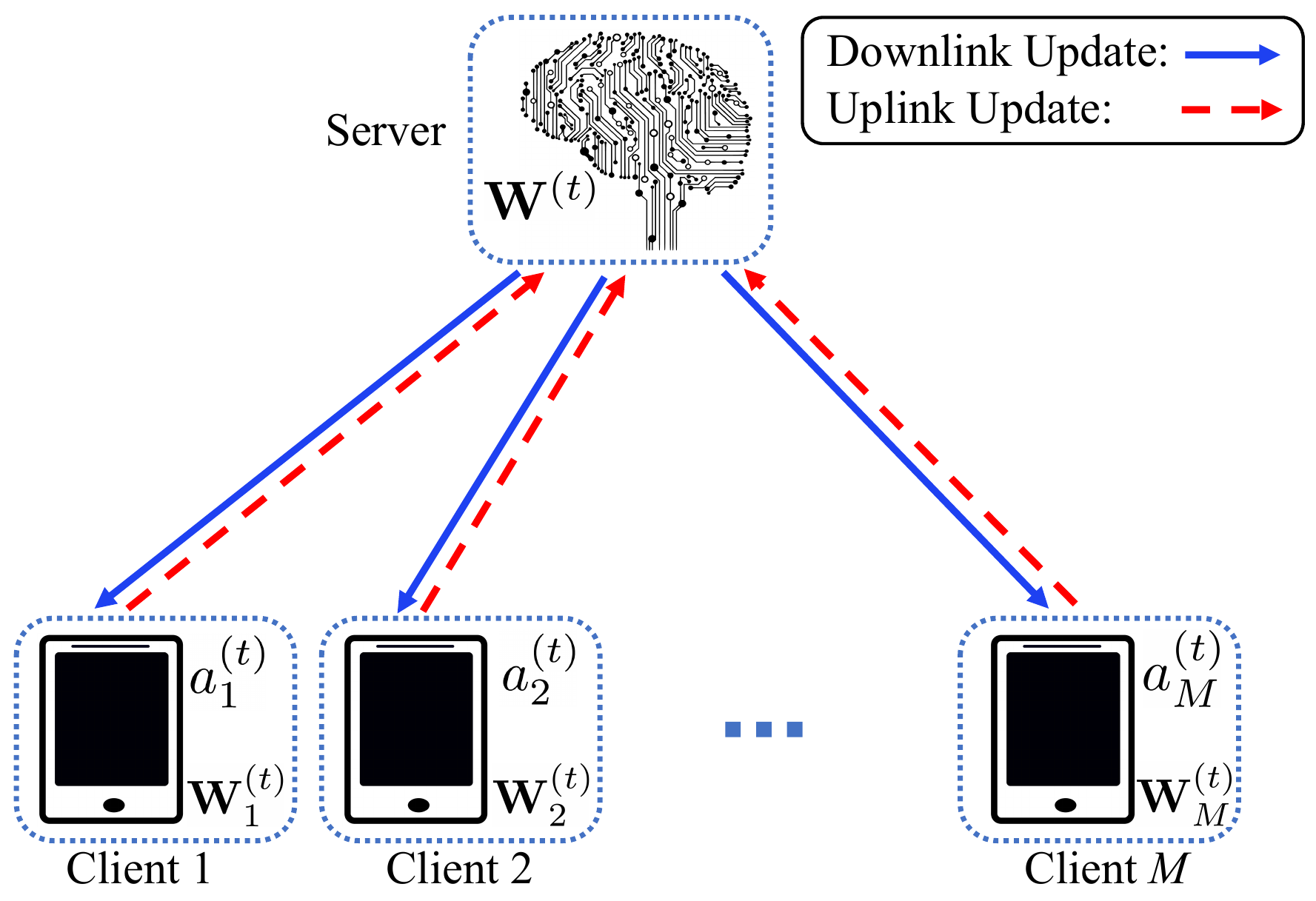}}
  
  \end{minipage}
  
  \caption{An illustration of the FL framework, with $M$ clients. In communication round $t$, the server selects and distributes a global model $\mathbf{W}^{(t)}$ to a subset of clients based on $\bf{a}^{(t)}$ (Downlink Update). Then, the selected clients compute and transmit their local models \{$\mathbf{W}^{(t)}_{i}$\} to the server (Uplink Update).}
  \label{fig:model}
  
  \end{figure}

\subsection{Preliminaries of FL and \texttt{FedAvg} Algorithm}
\label{ssec:subfed}

The FL framework consists of one central server and multiple clients. Clients participate in training a shared global model under the coordination of the server, without having to share private data. 
Given $M$ clients, let $n_k$ be the number of datapoints that client $k$ has, and let $n := n_1 + \dots + n_M$ be the total number of datapoints. In the usual FL set-up, the stochastic vector $\mathbf{n} := [\tfrac{n_1}{n}, \dots, \frac{n_M}{n}]$ represents the discrete probability distribution used for client selection, where $p_k := \tfrac{n_k}{n}$ is the probability that client $k$ is selected in each communication round. 
The optimization problem that FL tackles can thus be formulated as the minimization problem:
\vspace*{0.6em}
\[\min _{w} f(w)=\smash{\sum_{k=1}^{M}} \ p_{k} f_{k}(w),\]
where $f_{k}$ is the local loss function of client $k$, whose input $w$ is the set of model parameters of a fixed model architecture.  

The \texttt{FedAvg} algorithm summarizes how this FL optimization problem is solved~\cite{mcmahan2017communication}. 
In each communication round, a small group of clients is selected and local training is performed at each client, based on the same model downloaded from the global server. 
The main idea is that the gradient updates from the clients are aggregated at the server via a weighted average. 
At the end of round $t$, the server updates the model parameters with gradient descent
via $w_{t+1} \leftarrow w_{t}-\eta \nabla\!f(w_{t})$, 
where $\nabla\!f(w_{t})= \sum_{k \in \mathcal{S}_t} \tfrac{n_k}{n_{\mathcal{S}_t}}g_{k}$, 
and $\eta$ is the learning rate. 
Here, $g_{k}$ is the local update of selected client $k$, 
\smash{$\mathcal{S}_t$} is the subset of selected clients in round $t$, 
and \smash{$n_{\mathcal{S}_t} := \sum_{i\in \mathcal{S}_t} n_i$}. 
The number $K = |\mathcal{S}_t|$ of selected clients is calculated by the formula $K=\gamma \cdot M$, where $\gamma$ (satisfying $0<\gamma<1$) denotes the \textit{fraction} of the selected clients.

In this usual \texttt{FedAvg} algorithm, both the probability distribution for client selection and the fraction $\gamma$ of selected clients, are kept invariant throughout all communication rounds. 
In the next two subsections, we explain how \texttt{AdaFL} varies the client selection probability distribution, and varies the fraction $\gamma$, respectively.

\subsection{Attention Mechanism}
\label{ssec:subattention}

In any real-world FL implementation on non-IID client data, the relative training performances of different clients cannot be predicted in advance. Different clients could have different relative importance towards model aggregation, which could vary over different communication rounds. 
To avoid making idealistic assumptions, we shall take a data-driven approach. We introduce an attention mechanism to measure the relative importance of the different clients, and adjust the probability distribution for client selection accordingly, based on real-time local training performance. 
Our approach differs from existing FL approaches, e.g. ~\cite{konevcny2016federated2,mcmahan2017communication,li2019fair,ji2019learning}, where client selection is not modified and hence independent of the local training performance of the clients.

We shall use Euclidean distance as a measure of the model divergence of each local model, relative to the global model. 
The vector of attention scores $\mathbf{a}^{(t)}=[a^{(t)}_1, a^{(t)}_2,..., a^{(t)}_M]$ in round $t$ is identical to the corresponding client selection probability distribution $\mathbf{p}=[ p_1,p_2,...,p_M]$ for that round $t$, and it is initialized in the first round as $\mathbf{a}^{(1)}= \mathbf{n} $.

Specifically, at the beginning of round $t$, the server selects $K$ clients, according to the probability distribution $\mathbf{p}$.
Local training then occurs. We shall denote the local models of the selected clients by
\smash{$\mathbf{W}^{(t)}_{i_1}, \mathbf{W}^{(t)}_{i_2},...,\mathbf{W}^{(t)}_{i_K}$}, where $i_j$ is the index of the $j$-th client in the selected subset $\mathcal{S}_t$ (for round $t$).
Each $\mathbf{W}^{(t)}_{i_j}$ is a collection of weight matrices for the layers of the neural network.
After aggregation at the server, we obtain a new global model, denoted by $\mathbf{W}^{(t+1)}$. 
The process in a typical round is shown in Fig. \ref{fig:model}. 

Identify each $r$-by-$s$ weight matrix as a vector in $\mathbb{R}^{rs}$ and concatenate all such vectors, so that \smash{$\mathbf{W}^{(t)}_{i_j}$} is represented by a single parameter vector \smash{$\mathbf{w}^{(t)}_{i_j}$}. 
Thus, the $K$ local models and the new global model are represented by the vectors 
\smash{$\mathbf{w}^{(t)}_{i_1}, \mathbf{w}^{(t)}_{i_2},\dots ,\mathbf{w}^{(t)}_{i_K}$} and
\smash{$\mathbf{w}^{(t+1)}$} respectively. 
For selected client $i \in \mathcal{S}_t$ in round $t$, we calculate the Euclidean distance between the global and local parameter vectors as follows: 
\begin{equation}\label{eqn:dt}
d^{(t)}_{i}=\left\|\mathbf{w}^{(t+1)}-\mathbf{w}_{i}^{(t)}\right\|_{2}.
\end{equation}
To reduce the fluctuations in attention scores of consecutive rounds, we incorporate the current attention score in our updating criterion: 
\begin{equation}\label{eqn:at}
a^{(t+1)}_{i}=\alpha a^{(t)}_{i} + (1-\alpha) \cdot \frac{d^{(t)}_{i}}{\sum_{k \in \mathcal{S}_t}d^{(t)}_{k}} \sum_{k \in \mathcal{S}_t}a^{(t)}_{k},
\end{equation}
where $\alpha \in [0,1)$ represents a decay rate of previous attention score contributions.  
For an unselected client $j$, we set \smash{$a^{(t+1)}_j=a^{(t)}_j$}.  
Note that $\mathbf{a}^{(t+1)}$ remains as a stochastic vector. 
Client selection in round $t+1$ then follows the updated
probability distribution $\mathbf{p} = \mathbf{a}^{(t+1)}$. 

Since our work does not deal with the usual federated optimization, our attention mechanism only updates the client selection probability distribution and will not change the aggregation weights. Moreover, the additional communication cost brought by our proposed algorithm is negligible. 
Notice also that by \eqref{eqn:dt} and \eqref{eqn:at}, a larger Euclidean distance between the vectors for the global model \smash{$w^{(t+1)}$} and local model \smash{$w^{(t)}_{i}$} will increase the probability that client $i$ would be selected in the next communication round. Subsequently, more training and computation would be done at the corresponding clients to obtain a better performance.
Hence, the overall effect is a fairer scheme, whereby the variance of local training performances among the clients is deliberately reduced.

\subsection{Dynamic Fraction for  Client Selection }
\label{ssec:subada}

How do we choose a ``good'' fraction $\gamma$ for client selection? What is considered ``good'' will depend on how we want to balance the trade-off between communication efficiency and performance stability. 
A smaller $\gamma$ would mean a lower communication cost in each communication round, but at the expense of larger fluctuations in training performance, and a slower rate of convergence \cite{li2019convergence}; this implies that more training rounds are required to reach a desired performance level. 
A larger $\gamma$ would bring better performance stability, and less communication rounds, but at the expense of a larger total communication cost. 
Under the usual assumption in FL that $\gamma$ is fixed throughout training, we would then be forced to choose between communication efficiency and performance stability.

\begin{figure}[tb!]

  \begin{minipage}[b]{1.0\linewidth}
    \centering
    \centerline{\includegraphics[width=8.5cm]{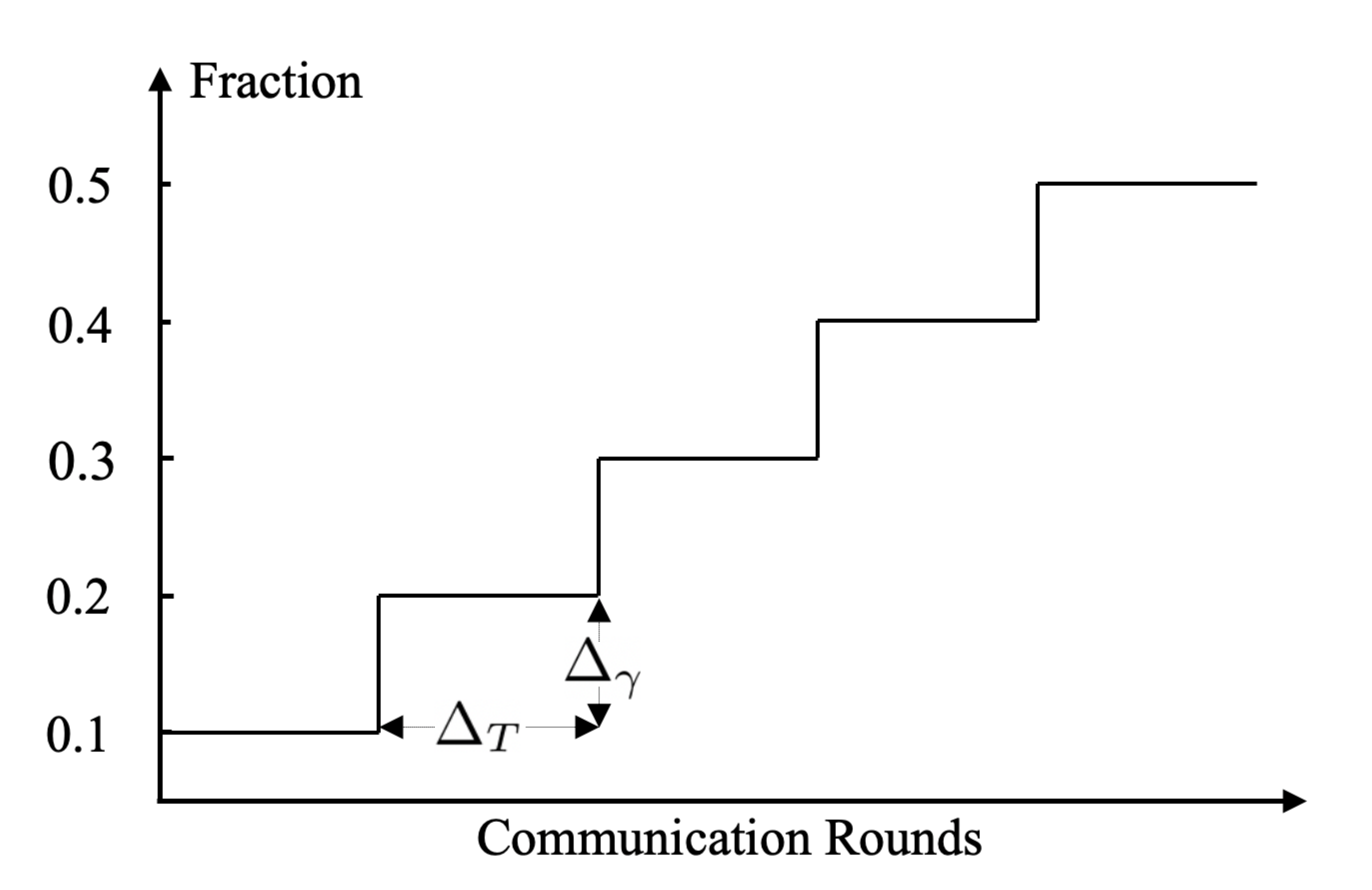}}
  \end{minipage}
  \caption{An example of our dynamic fraction scheme. $\Delta_T$  is the fixed 
  step size (number of communication rounds) between two consecutive fraction updates in the training. $\Delta_\gamma$ is the increasing step of the fraction.}
  \label{fig:ada_frac}
  \end{figure}

To circumvent this trade-off, we drop the assumption that $\gamma$ is fixed, and propose a dynamic fraction method, which adopts different fractions during different training stages, with the fraction increasing progressively. As an explicit example, depicted in Fig. \ref{fig:ada_frac}, we begin training with a small fraction $0.1$, and end training with a large fraction $0.5$. (Here, $0.1$ and $0.5$ are arbitrarily selected.) When using gradient descent, our method yields a relatively good performance, even with fewer clients involved at the beginning of training. With an increased amount of local training data in subsequent rounds, the training performance would have a more stable convergence. Intuitively, the updated global model gradients would be closer to optimal gradients that reflect the true data distribution, as there are increasingly more clients (and hence more data) involved in training, as the fraction increases.

To represent dynamic fractions, we shall use $\boldsymbol{\gamma}$ to denote the vector of fractions, whose $t$-th entry $\boldsymbol{\gamma}^{(t)}$ is the fraction used in the $t$-th communication round. 
Since $T$ is the total number of communication rounds, $\boldsymbol{\gamma}^{(1)}$ (resp. $\boldsymbol{\gamma}^{(T)}$) is the starting (resp. ending) fraction used in our dynamic fraction method. 
For simplicity, we recommend using a fixed step size $\Delta_T$ between consecutive fraction updates, and a fixed increment $\Delta_{\gamma}$ when updating the fraction $\gamma$; this means that $\Delta_T=\tfrac{1}{F} \cdot T$ and $\boldsymbol{\gamma}^{(T)} - \boldsymbol{\gamma}^{(1)} = (F-1)\Delta_{\gamma}$, where $F$ is the desired number of distinct fractions to be used. The vector $\boldsymbol{\gamma}$ can be computed accordingly. 
Observe that in our running example (see Fig.~\ref{fig:ada_frac}), we used $F = 5$, so that each fraction update, which increases the value of $\gamma$ by $\Delta_{\gamma} = \tfrac{1}{F} = 0.1$, occurs at every $\Delta_T = \tfrac{1}{F}\cdot T = 0.2T$ communication rounds. 

Although we only consider fixed $\Delta_T$ and $\Delta_\gamma$ for simplicity, it should be noted that our method works for any number $F$ of fraction values and any non-constant $\Delta_{\gamma}, \Delta_T$, 
and more generally, for any non-constant fraction $\gamma$ that is monotonically increasing from $\boldsymbol{\gamma}^{(1)}$ to $\boldsymbol{\gamma}^{(T)}$. 
In this paper, we do not address all the different (infinitely many) types of monotonically increasing $\gamma$ that could be used for our proposed dynamic fraction method; In our experiments, this simple use of multiple fraction values as described above, is already sufficient to yield performance improvement over the use of constant fraction.

\subsection{Algorithm Summary}
\label{ssec:subalgo}
Adaptive Federated Learning (\texttt{AdaFL}), our proposed method, combines both the attention mechanism described in Section~\ref{ssec:subattention}, and the dynamic fraction method described in Section~\ref{ssec:subada}.
The juxtaposition of these two components, on top the usual \texttt{FedAvg} algorithm, incorporates adaptive training adjustments, thereby yielding better communication efficiency with better performance stability. 
We give an overview of \texttt{AdaFL} in Algorithm 1.
\begin{algorithm}
  \caption{Adaptive Federated Learning \texttt{(AdaFL)}}
  \label{alg: AdaFL}
  \textbf{Inputs:} $M, T, \boldsymbol{\gamma}, \alpha, \mathbf{W}^{(1)}, \mathbf{n}$
  \begin{algorithmic}[1] 
\STATE{$\mathbf{a}^{(1)} \gets \mathbf{n}$} 
\FOR {$t=1$ \textbf{to} $T$}
    \STATE{$\mathbf{p} \gets \mathbf{a}^{(t)}$ and $K \gets \boldsymbol{\gamma}^{(t)} \cdot M$}
    \STATE{Server selects a subset of clients $\mathcal{S}_t$ of size $|\mathcal{S}_t|=K$}
		\STATEx{\ \ \ \ using probability distribution $\mathbf{p}$}
    \STATEx{\vspace*{-0.7em}}
    \STATEx{\ \ \ \ \textit{// local computation at clients}}
    \FOR{\textbf{selected client} $k \in \mathcal{S}_t$}
        \STATE{Client $k$ downloads global model $\mathbf{W}^{(t)}$}
        \STATE{Client $k$ computes local model $\mathbf{W}_k^{(t)}$}
    \ENDFOR
    \STATEx{\vspace*{-1.0em}}
    \STATEx{\ \ \ \ \textit{// global computation at server}}
    \STATE{Server computes a new global model by aggregation:
    \vspace*{-0.3em}
    \[\mathbf{W}^{(t+1)} \leftarrow \textstyle\sum_{k \in \mathcal{S}_t}  \tfrac{n_k}{n_{\mathcal{S}_t}}\mathbf{W}^{(t)}_{k}\]}
    \vspace*{-1.0em}
    \FOR {\textbf{selected client} $i \in \mathcal{S}_t$}
        \STATE{Server updates $d^{(t)}_{i} \gets \big\|\mathbf{w}^{(t+1)}-\mathbf{w}_{i}^{(t)}\big\|_{2}$ \ \ and}
				\STATEx{\ \ \ \ \ \ \ \ $a^{(t+1)}_{i} \gets \alpha a^{(t)}_{i} + (1-\alpha) \cdot \tfrac{d^{(t)}_{i}}{\sum_{k \in \mathcal{S}_t}d^{(t)}_{k}} \textstyle\sum_{k \in \mathcal{S}_t}a^{(t)}_{k}$}
    \ENDFOR
    \FOR {\textbf{unselected client} $j \notin \mathcal{S}_t$}
        \STATE{Server updates $a^{(t+1)}_j \gets a^{(t)}_j$}
    \ENDFOR
\ENDFOR
  \end{algorithmic}

\end{algorithm}
The key difference of our proposed \texttt{AdaFL} algorithm over the \texttt{FedAvg} algorithm is the adaptive parameter adjustment scheme. 
Parameters are adjusted by using real-time information from local training. 
Hence, the central server also plays a dual role as a resource allocator during training, in addition to its usual role of coordinating the aggregation of local weights in each communication round. 
The resource allocation can be made fairer by improving the weights of clients with larger model divergence. 

It should be noted that our proposed algorithm complements existing communication-efficient federated algorithms, such as compression~\cite{konevcny2016federated2,sattler2019robust}, data augmentation~\cite{yoon2021fedmix} and some optimization methods for federated learning~\cite{li2020prox,karimireddy2020scaffold}. 
Later in our experiments, we show how our proposed \texttt{AdaFL} can enhance the performance of existing popular FL algorithms. 

\section{Experiments}
\label{sec:exp}

In this section, we describe the details of our experiments, and evaluate the performance of our proposed \texttt{AdaFL} algorithm.

\subsection{Experiment Setup}
\label{ssec:subexp}

We evaluate our \texttt{AdaFL} algorithm on image classification tasks on two image datasets, MNIST~\cite{lecun1998gradient} 
and CIFAR-10~\cite{krizhevsky2009learning}, with neural network models. 
For experiments on the MNIST dataset, we train a Multi-Layer-Perceptron (MLP) model (2 hidden layers, each layer with 200 units and ReLU activation) with a fixed learning rate $\eta=0.01$. For training data samples, we use the non-IID data partition as described in \cite{mcmahan2017communication}. 
For experiments on the CIFAR-10 dataset, we train a CNN model with the same model architecture as given in \cite{yoon2021fedmix}, with IID data partition. We used an initial learning rate of $\eta=0.01$ with a decay of $0.99$ across the communication rounds. 
In all our experiments, across all FL algorithms, we train a local model at each selected client via stochastic gradient descent (SGD), using momentum coefficient $0.5$.

The remaining parameter settings are given as follows. We use $M=100$ clients for training. The number of local epochs and batch size are set as $E=5$ and $B=10$ respectively. 
The initial attention score vector $\mathbf{a}^{(1)}$ is determined by the local dataset size as described in Algorithm \ref{alg: AdaFL}. 
For our attention score update process, we fix $\alpha=0.9$. 
For the local dataset size, we consider a balanced data distribution for all experiments, in which every client has the same local dataset size (number of local training samples); this implies that the initial attention score of each client is $\frac{1}{100}$. 
For the dynamic selection of fractions, we use $\boldsymbol{\gamma}^{(1)}=0.1$ and $\boldsymbol{\gamma}^{(T)}=0.5$, with increments of $\Delta_\gamma=0.1$ (see Fig. \ref{fig:ada_frac}). 

\subsection{Ablation Study}
\label{ssec:subcomp}

\begin{figure}[tb!]

    \begin{minipage}[b]{1.0\linewidth}
      \centering
      \centerline{\includegraphics[clip,trim=0 0.2cm 0 0.95cm,width=9.5cm]{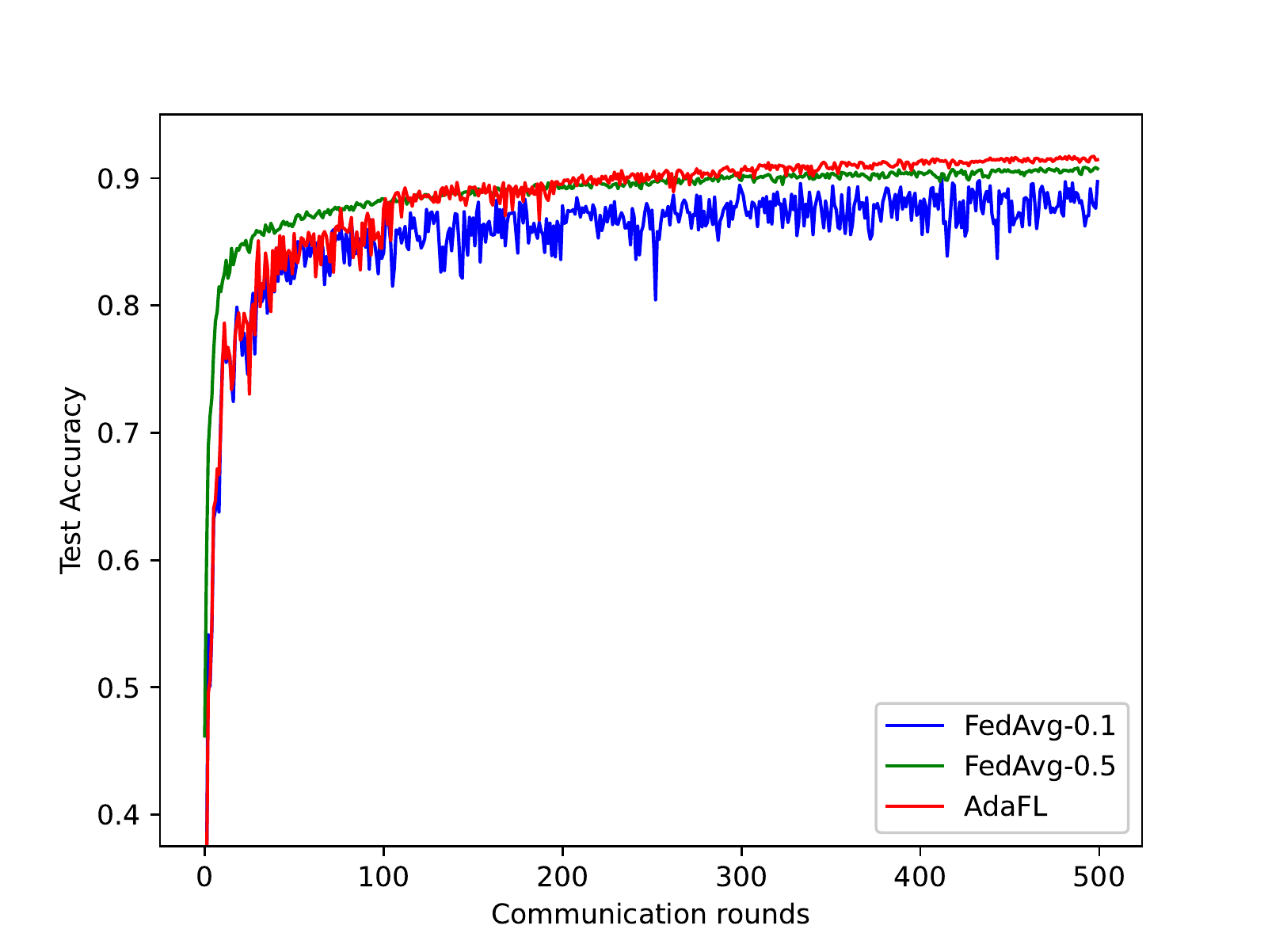}} 
    
    \end{minipage}
    
    \caption{A comparison of the test accuracy between \texttt{AdaFL} and \texttt{FedAvg} on MNIST with non-IID local data. 
    Here, we used a dynamic fraction scheme, starting from $\gamma = 0.1$ and increasing to $\gamma = 0.5$. 
    This figure shows that (i): \texttt{AdaFL} has better performance stability than \texttt{FedAvg} at smaller fractions; and (ii): \texttt{AdaFL} has better test accuracy than \texttt{FedAvg} at larger fractions.}
    \label{fig:acc}
    
    \end{figure}

We report our ablation study results in Tables \ref{table:ablation} and \ref{table:rnd_ablation}, in which we evaluate the performances of our proposed \texttt{AdaFL} on the two datasets, with and without the two components in \texttt{AdaFL}: the attention mechanism and the dynamic fraction method. 
We use \texttt{FedAvg-0.1} and \texttt{FedAvg-0.5} as our baselines, which refer to the usual \texttt{FedAvg} algorithm with a constant fraction of $\gamma = 0.1$ and $\gamma = 0.5$ respectively. 
For illustration, a comparison of \texttt{AdaFL} with both baselines \texttt{FedAvg-0.1} and \texttt{FedAvg-0.5}, over all $500$ communication rounds of our experiments on MNIST, is given in Fig. \ref{fig:acc}. In particular, Fig. \ref{fig:acc} shows that \texttt{AdaFL} has the advantages of both \texttt{FedAvg-0.1} and \texttt{FedAvg-0.5}, which starts training with smaller communication cost, and ends training with better performance stability and test accuracy. 

For both Tables \ref{table:ablation} and \ref{table:rnd_ablation}, we write \texttt{Attn.-0.1} and \texttt{Attn.-0.5} to mean that we apply only the attention mechanism to \texttt{FedAvg} with a constant fraction of $\gamma = 0.1$ and $\gamma = 0.5$ respectively, while we write \texttt{Dyn.\,FedAvg} to mean that we apply only the dynamic fraction method (with the fraction increasing progressively from $0.1$ to $0.5$, with $T=1500$) to \texttt{FedAvg}, without the attention mechanism.

In Table \ref{table:ablation}, we report the ``best test accuracies'' of all aforementioned methods, on both datasets. However, it should be noted that our chosen model architectures do not yield state-of-the-art ``best accuracies'' on the respective datasets, 
since our goal is to show that \texttt{FedAvg} can be improved with our two proposed components in \texttt{AdaFL}. 
Due to the natural random fluctuations in test accuracy performance over consecutive communication rounds, we shall also report ``average test accuracy'' as a measure of performance stability; see Section \ref{sssec:ablacc} for more details. 
In Table \ref{table:rnd_ablation}, we report the number of communication rounds and total communication cost that each method takes to reach a specified target accuracy,  
which is chosen to be close to the corresponding best accuracy as given in Table \ref{table:ablation}; see Section \ref{sssec:ablcomm} for more details.

For the rest of this subsection, we shall evaluate the two components of \texttt{AdaFL} (using $T=1000$) with respect to (i) accuracy and performance stability; and (ii) required number of communication rounds and total communication cost, respectively.

\subsubsection{Accuracy Performance}
\label{sssec:ablacc}    
We use average accuracy and best accuracy to evaluate the outperformance and convergence stability. To better capture the notion of performance stability, we use the average accuracy of the last $\ell$ rounds as a key performance metric (we use $\ell = 10$ in our experiments). As the results in Table \ref{table:ablation} show, our experiments on both datasets, for which training end with a larger fraction (\texttt{AdaFL}, \texttt{Attn.-0.5} and \texttt{Dyn.\,FedAvg}) have higher average accuracies and hence better performance stability.  

Thus, the attention component increases model accuracy, while the dynamic fraction component improves performance stability. 
Compared to the two baselines (\texttt{FedAvg-0.1} and \texttt{FedAvg-0.5}), \texttt{AdaFL} achieved higher accuracies  
on both MNIST ($0.43\%$--$2.45\%$) and CIFAR-10 ($0.71\%$--$1.5\%$), with better performance stability. 
The experiments with the attention mechanism incorporated (\texttt{AdaFL}, \texttt{Attn.-0.1} and \texttt{Attn.-0.5}) have higher best accuracy performance. 

\begin{table}[htb]
    
    \centering
    \begin{tabular}{>{\small}ccccc}
    
    \toprule
    
    \multirow{2}{*}{\bf{Algorithm}} 
    & \multicolumn{2}{c}{\bf{MNIST}} & \multicolumn{2}{c}{\bf{CIFAR-10}} \\
    \cmidrule(){2-3} \cmidrule(){4-5}
    
                  & Average   & Best       & Average   & Best        \\ \midrule
    \addlinespace[0.04cm]
  
    \texttt{AdaFL}  & {\bf 91.13}& {\bf 91.64}&  74.38   & {\bf 76.17}         \\ 
    \addlinespace[0.01cm]
    \hline
    \addlinespace[0.04cm]
    \texttt{Attn.-0.1}  &  88.92     &  91.30     & 73.13       & 74.91             \\
    \addlinespace[0.04cm]
    \texttt{Attn.-0.5}  &  91.07    &  91.58     &  {\bf74.42}  & 75.96             \\
    \addlinespace[0.04cm]
    \texttt{Dyn.\,FedAvg}   &90.33     &91.19      & 74.33       & 75.04             \\
    \addlinespace[0.04cm]
    \hline
    \addlinespace[0.04cm]
    \texttt{FedAvg-0.1}    & 88.68    & 91.05    & 72.88    &  74.82     \\ 
    \addlinespace[0.04cm]
  
    \texttt{FedAvg-0.5}    & 90.40    & 91.21    & 73.67     &  75.31       \\

    \addlinespace[0.01cm]
  
    \bottomrule
    \end{tabular}
    \caption{Ablation study in terms of average and best test accuracy performances (averaged over $3$ trials). Each value in the Average column represents the average test accuracy of the last $10$ communication rounds, which serves dually as a measure of performance stability. Each value in the Best column represents the best test accuracy achieved over all communication rounds.}
    \label{table:ablation}
    \end{table}

\subsubsection{Communication Efficiency Performance}
\label{sssec:ablcomm} 

We define communication cost in terms of relative units, where each relative unit is the cost of transmitting data representing all parameter updates of a single neural network model, from a single client to the global server, in a single communication round. 
Given $\boldsymbol{\gamma}$ and the required number $T^*$ of communication rounds to reach target accuracy, we define the \textit{total communication cost} to be {$\sum_{t=1}^{T^*} \boldsymbol{\gamma}^{(t)} \cdot M$}; this value is the total number of relative units across all $T^*$ communication rounds. 
Hence, \textit{better communication efficiency} shall mean a lower total communication cost. 

\begin{table}[htb]
    
    \centering
    \begin{adjustbox}{width=\columnwidth,center}
  
    \begin{tabular}{ccccc}
   
    \toprule
     
    \multirow{2}{*}{\bf{Algorithm}} 
      & \multicolumn{2}{c}{\bf{MNIST}} & \multicolumn{2}{c}{\bf{CIFAR-10}} \\
    \cmidrule(){2-3} \cmidrule(){4-5}
    
                  & 90\%      & 91\%   &      &  73\%      \\ \midrule
    \addlinespace[0.01cm]
    
    \texttt{AdaFL}  & {\bf 423}\,({\bf 6690}) &{\bf761}\,({\bf18440})& &{\bf 683}\,({\bf 15320})    \\ 
    \addlinespace[0.04cm]
    \hline
    \addlinespace[0.04cm]
    \texttt{Attn.-0.1} &939\,(9390) &1952\,(19520) & & 1571\,(15710) \\
    \addlinespace[0.04cm]
    \texttt{Attn.-0.5} &420\,(21000) &741\,(37050) & &635\,(31570) \\
    \addlinespace[0.04cm]
    \texttt{Dyn.\,FedAvg} &951\,(20040) & 1485\,(44250)& &1103\,(26120) \\
    \addlinespace[0.04cm]
    \hline
    \addlinespace[0.04cm]
    \texttt{FedAvg-0.1}    & 1008\,(10080)   &  2528\,(25280)  &       &   1957\,(19570)    \\ 
    \addlinespace[0.04cm]
    
    \texttt{FedAvg-0.5}    & 570\,(28500)  &  1232\,(61600)   &       &   892\,(44600)    \\ 
    \addlinespace[0.01cm]
    
    \bottomrule
    \end{tabular}
    \end{adjustbox}
    \caption{Ablation study in terms of the required number of rounds (values preceding brackets) and the total communication cost (values in brackets) to reach the specified target test accuracy. All reported values are averaged over $3$ trials. For a fair evaluation, the stopping criterion used is that the average test accuracy of the last $5$ rounds must exceed the target test accuracy.}
    \label{table:rnd_ablation}
    
    \end{table}

As Table \ref{table:rnd_ablation} shows, the use of larger fractions would require less communication rounds to reach the specified target accuracies, while the use of small fraction (with lower communication cost per round) would require more communication rounds to reach stable convergence, i.e. a larger total communication cost. 
In comparison, the use of dynamic fractions not only yields faster stable convergence, but also has better communication efficiency. 

To conclude, the ablation study results reported in Tables \ref{table:ablation} and \ref{table:rnd_ablation} show conclusively that both components in \texttt{AdaFL} contribute towards the outperformance over \texttt{FedAvg}.

\subsection{Performance Evaluation}
\label{perf_ana}
As discussed earlier in Section~\ref{sec:intro}, 
\texttt{FedProx} and \texttt{SCAFFOLD} are federated optimization methods, while \texttt{FedMix} is a data augmentation method designed for FL. 
These algorithms employ a fixed probability distribution for client selection and a fixed fraction $\gamma$ throughout training. 
In this subsection, we report how our proposed \texttt{AdaFL} can be incorporated to further improve these algorithms. 
Table \ref{table:compare_acc} shows that the incorporation of \texttt{AdaFL} improves both model accuracy and performance stability, while Table \ref{table:compare_rnds} shows that the incorporation of \texttt{AdaFL} reduces the required number of communication rounds and total communication cost to reach target accuracy.

\begin{table}[htb]
    
    \centering
    \begin{tabular}{>{\small}ccccc}
   
    \toprule
    
    \multirow{2}{*}{\bf{Algorithm}} 
    & \multicolumn{2}{c}{\bf{MNIST}} & \multicolumn{2}{c}{\bf{CIFAR-10}} \\
    \cmidrule(){2-3} \cmidrule(){4-5}
    
                  & Average   & Best       & Average   & Best        \\ \midrule
    \addlinespace[0.04cm]
  
    \texttt{AdaFL+FedProx}   & {\bf91.67}   &  {\bf92.42}    &   {\bf74.94}       &  {\bf76.24}          \\
    \addlinespace[0.04cm]
  
    \texttt{FedProx-0.1}   &   89.15      &    91.46       &  72.88       &  75.90           \\
    \addlinespace[0.04cm]
  
    \texttt{FedProx-0.5}   &   90.81     &     91.55      &   73.57       &  76.12           \\
    \addlinespace[0.04cm]
    \hline
    \addlinespace[0.04cm]
    \texttt{AdaFL+FedMix}   & {\bf90.52}    & {\bf91.30}  &{\bf73.27}    &   {\bf75.05}          \\
    \addlinespace[0.04cm]
  
    \texttt{FedMix-0.1}   & 88.37      &  90.61       &  71.53        &   73.43          \\
    \addlinespace[0.04cm]
  
    \texttt{FedMix-0.5}   &  89.91     &  91.08        & 72.42         &  74.12           \\
  
    \addlinespace[0.04cm]
    \hline
    \addlinespace[0.04cm]
    \texttt{AdaFL+SCAFFOLD}   & {\bf90.30}   & {\bf91.52}        & {\bf74.98}       &    {\bf75.53}         \\
    \addlinespace[0.04cm]
  
    \texttt{SCAFFOLD-0.1}   &  87.82      &  89.96      &  71.62      &     74.12        \\
    \addlinespace[0.04cm]
  
    \texttt{SCAFFOLD-0.5}   &   89.73      &  90.82     &  73.50      &   74.77       \\
  
    \addlinespace[0.01cm]
  
    \bottomrule
    \end{tabular}
    \caption{A comparison of the average and best test accuracy performance for various FL algorithms (averaged over $3$ trials). \texttt{AdaFL+FedProx} stands for \texttt{AdaFL} incorporated into \texttt{FedProx}, while \texttt{FedProx-0.1/0.5} refers to the usual \texttt{FedProx} with constant fraction $0.1$ or $0.5$. The other algorithms reported in this table are defined analogously. We use the same performance metrics as used in Table \ref{table:ablation}. In particular, average test accuracy (of the last $10$ communication rounds) is a measure of performance stability.}
    \label{table:compare_acc}
    \end{table}

\begin{table}[htb]
    
    \centering

    \begin{tabular}{>{\small}c>{\small}c>{\small}c}
    
    \toprule
    
     {\bf{Algorithm}}   & {\bf{MNIST}}     &{\bf{CIFAR-10}} \\\midrule
     
    \addlinespace[0.06cm]
                                & 91\%         &  73\%      \\
                            \cmidrule(){2-3}
    \addlinespace[0.01cm]
    
    \texttt{AdaFL+FedProx}   &{\bf 821\,(21600)} &721\,{\bf(16840)}                     \\
    \addlinespace[0.04cm]
  
    \texttt{FedProx-0.1}   &   2439\,(24390)  &   1762\,(17620)                     \\
    \addlinespace[0.04cm]
  
    \texttt{FedProx-0.5}   &   1084\,(54200)    &   {\bf658}\,(32900)                      \\
    \addlinespace[0.04cm]
    \hline
    \addlinespace[0.04cm]
                          & 90\%         &  72\%      \\
                            \cmidrule(){2-3}
    \addlinespace[0.01cm]
    \texttt{AdaFL+FedMix}   & {\bf  852\,(22600)}   &  {\bf  698\,(15920) }                   \\
    \addlinespace[0.04cm]
  
    \texttt{FedMix-0.1}   &    2275\,(22750)   &   1903\,(19030)                       \\
    \addlinespace[0.04cm]
  
    \texttt{FedMix-0.5}   &    1241\,(62050)   &   732\,(36600)                     \\
  
    \addlinespace[0.04cm]
    \hline
    \addlinespace[0.04cm]
                         & 89\%         &  72\%      \\
                            \cmidrule(){2-3}
    \addlinespace[0.01cm]
    \texttt{AdaFL+SCAFFOLD}   &  {\bf 794\,(19760) }        &  {\bf  672\,(15600)  }        \\
    \addlinespace[0.04cm]
  
    \texttt{SCAFFOLD-0.1}   &    2252\,(22520)  &     1981\,(19810)            \\
    \addlinespace[0.04cm]
  
    \texttt{SCAFFOLD-0.5}   &     1034\,(51700) &    725\,(36250)          \\
    \addlinespace[0.01cm]
    
    \bottomrule
    \end{tabular}
    \caption{A comparison of the required number of communication rounds (values preceding brackets) and total communication cost (values in brackets) to reach target test accuracy, for various algorithms (averaged over $3$ trials). Similar to the settings used in Table~\ref{table:rnd_ablation}, the target test accuracies are chosen based on the results of Table~\ref{table:compare_acc}. For a fair evaluation, the stopping criterion used is that the average accuracy of last $5$ rounds must exceed the target accuracy.}
    \label{table:compare_rnds}
    
    \end{table}
  
Overall, \texttt{AdaFL} complements the performance of the three state-of-the-art algorithms on both datasets (see boldfaced values in Tables \ref{table:compare_acc} and \ref{table:compare_rnds}), with respect to test accuracy and communication efficiency. 
For test accuracy, the \texttt{AdaFL}-based experiments yield better performance, with improvements on both MNIST (increase of $2.52\%$, $2.15\%$, and $2.48\%$ respectively) and CIFAR-10 (increase of $2.06\%$, $1.74\%$, and $3.06\%$ respectively), for all three algorithms. 
Also, observe that our \texttt{AdaFL} requires the least number of communication rounds in most of the experiments and has the lowest total communication cost to reach the specified target accuracy for all the experiments, giving a $0.7\%$ to $21.3\%$ reduction in total communication cost for small fractions, and more significantly, a $48.8\%$ to $63.6\%$ reduction for large fractions. 

These results show conclusively that the incorporation of \texttt{AdaFL} into these state-of-the-art FL algorithms would enhance the performance of all three aspects: 
model accuracy, performance stability, and communication efficiency.

\section{Conclusion}
\label{sec:conc}
In this paper, we propose an attention-based federated learning algorithm with dynamic fraction for client selection, which we call \texttt{AdaFL}. 
It is a simple algorithm that can be easily incorporated into various state-of-the-art FL algorithms to obtain improvements on several aspects: model accuracy, performance stability, and communication efficiency.

Our detailed ablation study shows that the two components in \texttt{AdaFL} indeed contribute significantly towards the outperformance over the usual \texttt{FedAvg} algorithm, with respect to all three aspects.  
When incorporated into existing state-of-the-art FL algorithms, \texttt{AdaFL} yields consistently better performance. 
We foresee that \texttt{AdaFL} can easily be incorporated into subsequent FL algorithms, to enhance the performance, especially in non-IID data settings.

Our proposed \texttt{AdaFL} algorithm gives clients who have larger model divergence a higher chance to participate in training. Can \texttt{AdaFL} be used as a stepping stone to develop other methods to improve fairness in FL? We plan to further explore this issue. 
We also plan to study the attention mechanism in \texttt{AdaFL}, in the context of imbalanced data.

\section*{Acknowledgments}
This research is supported by the National Research Foundation, Singapore under its AI Singapore Programme (AISG Award No: AISG-RP-2019-015).

\bibliographystyle{named}
\bibliography{ijcai21}

\end{document}